%% file: main.tex
\documentclass[10pt,twocolumn,letterpaper]{article}

\usepackage{arxiv_iccv}
\usepackage{times}
\usepackage{epsfig}
\usepackage{graphicx}
\usepackage{amsmath}
\usepackage{amssymb}

\usepackage{booktabs}
\usepackage{varwidth}
\usepackage{placeins}
\usepackage{colortbl}

\usepackage[pagebackref=true,breaklinks=true,letterpaper=true,colorlinks,bookmarks=false]{hyperref}

\input{defines}
\input{macros}

\DeclareMathOperator*{\argmax}{\arg\!\max}

\newcommand{\ApproachName}{S2VT\xspace}
\iccvfinalcopy 

\def\iccvPaperID{1211} 

\ificcvfinal\fi
\begin{document}

\title{Sequence to Sequence -- Video to Text}
\author{Subhashini Venugopalan$^1$
\and Marcus Rohrbach$^{2,4}$
\and Jeff Donahue$^2$
\and  Raymond Mooney$^1$
\and Trevor Darrell$^{2}$
\and Kate Saenko$^3$
}
\date{
\begin{tabular}{cc}
$^1$University of Texas at Austin &$^2$University of California, Berkeley \\
$^3$University of Massachusetts, Lowell &  $^4$International Computer Science Institute, Berkeley\\
\end{tabular}
}

\maketitle

\begin{abstract}

Real-world videos often have complex dynamics; and methods for generating
open-domain video descriptions should be sensitive to temporal structure and allow both input (sequence of frames) and output (sequence of words) of variable length. 
To approach this problem, we propose a novel end-to-end sequence-to-sequence model to generate captions for videos. For this we exploit recurrent neural networks, specifically LSTMs, which have demonstrated state-of-the-art performance in image caption generation. 
Our LSTM model is trained on video-sentence pairs and learns to associate a sequence of video frames to a sequence of words in order to generate a description of the event in the video clip. Our model naturally is able to learn the temporal structure of the sequence of frames as well as the sequence model of the generated sentences, \ie a language model. 
We evaluate several variants of our model that exploit different visual features on a standard set of YouTube videos and two movie description datasets (M-VAD and MPII-MD). 

\end{abstract}

\input{intro}

\input{related}

\input{approach}

\input{evaluation}

\input{discussion}
\input{conclusion}
\input{ack}

\FloatBarrier
\small
\bibliographystyle{ieee}
\bibliography{biblioShort,rohrbach,related,refs}

\end{document}

%% file: defines.tex
 \makeatletter
 \DeclareRobustCommand\onedot{\futurelet\@let@token\@onedot}
 \def\@onedot{\ifx\@let@token.\else.\null\fi\xspace}
  
 \def\ie{i.e\onedot}

 \makeatother

\DeclareRobustCommand{\Tableref}[1]{Table~\ref{#1}}

\DeclareRobustCommand{\Tablesref}[1]{Tables~\ref{#1}}

%% file: macros.tex
\graphicspath{{./fig/}{./fig/plots/}}

%% file: intro.tex
\section{Introduction}

Describing visual content with natural language text has recently received increased interest, especially describing images with a single sentence \cite{donahue15cvpr,chen14arxiv,karpathy14arxiv,kiros14arxiv,kuznetsova14tacl,mao14arXiv,rohrbach13iccv,vinyals14arxiv}. Video description has so far seen less attention despite its important applications in human-robot interaction, video indexing, and describing movies for the blind.
While image description handles a variable length output sequence of words, video description also has to handle a variable length input sequence of frames. Related approaches to video description have resolved variable length input by holistic video representations \cite{rohrbach13iccv,rohrbach15cvpr,guadarrama13iccv}, pooling over frames \cite{venugopalan15naacl}, or sub-sampling on a fixed number of input frames \cite{yao15arxiv}.
In contrast, in this work we propose a sequence to sequence model which is trained end-to-end and is able to learn arbitrary temporal structure in the input sequence. Our model is sequence to sequence in a sense that it reads in frames sequentially and outputs words sequentially.
\begin{figure}[t]
\begin{center}
\includegraphics[width=\linewidth]{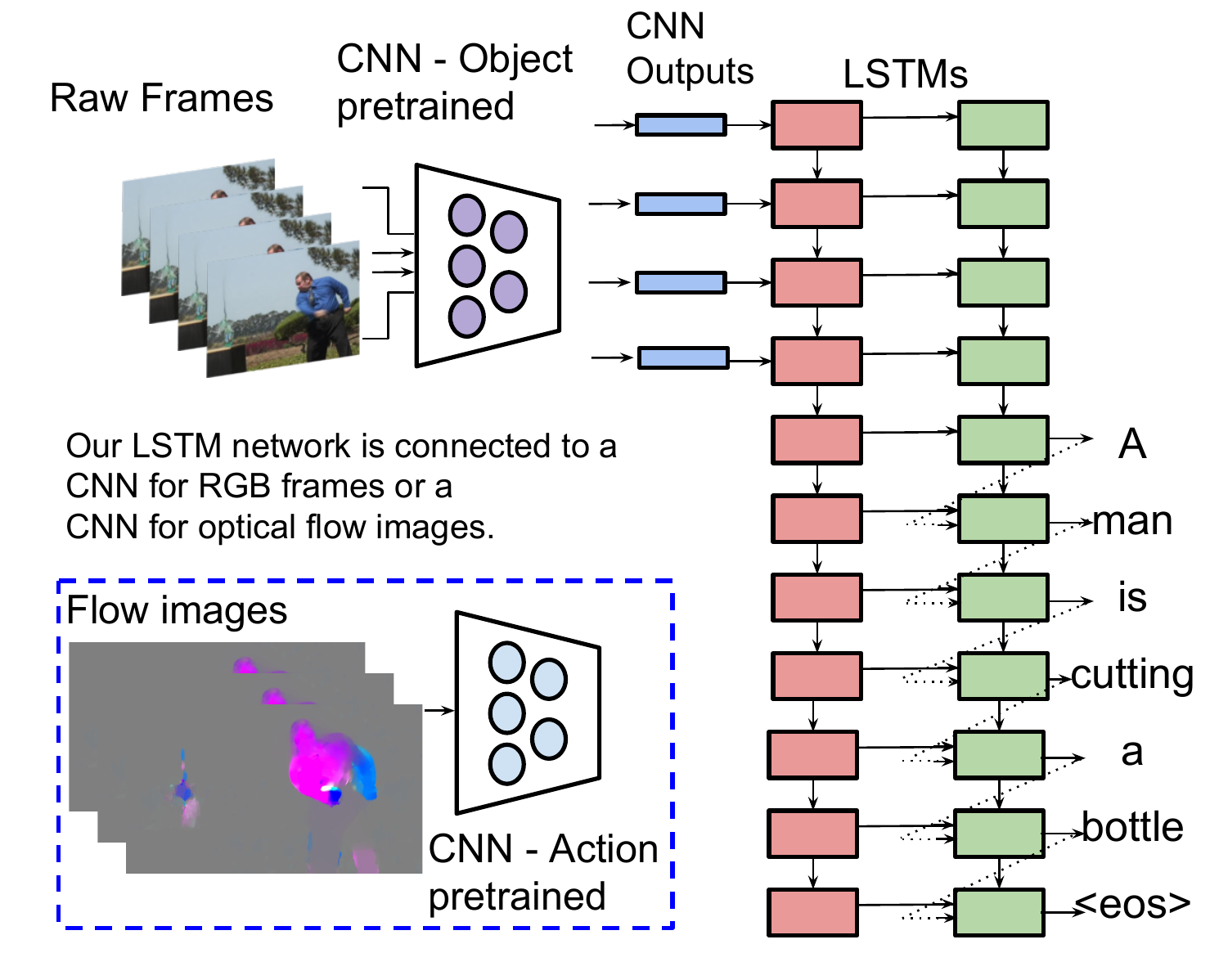}
\end{center}
 \caption{
Our  \ApproachName approach performs video description using a sequence to sequence model. It incorporates a stacked LSTM which first reads the sequence of frames and then generates a sequence of words. The input visual sequence to the model is comprised of RGB and/or optical flow CNN outputs.
}
\label{fig:teaser}
\end{figure}

The problem of generating descriptions in open domain videos is difficult not just due to the diverse set of objects, scenes, actions, and their attributes, but also because it is hard to determine the salient content and describe the event appropriately in context.  
To learn what is worth describing, 
our model learns from video clips and paired sentences that describe the depicted events in natural language.
We use Long Short Term Memory (LSTM) networks \cite{schmidLSTM}, a type of recurrent neural network (RNN) that has achieved great success on similar sequence-to-sequence tasks such as speech recognition~\cite{graves2014towards} and machine translation~\cite{ilya:nips14}. Due to the inherent sequential nature of videos and language, LSTMs are well-suited for generating descriptions of events in videos.

The main contribution of this work is to propose a novel model, S2VT,
which learns to directly map a sequence of frames to a sequence of words.
Figure \ref{fig:teaser} depicts our model.
A stacked LSTM first encodes the frames one by one, taking as input the output of a Convolutional Neural Network (CNN) applied to each input frame's intensity values. Once all frames are read, the model generates a sentence word by word. The encoding and decoding of the frame and word representations are learned jointly from a parallel corpus.
To  model the temporal aspects of activities typically shown in videos, we also compute the optical flow \cite{brox04flow} between pairs of consecutive frames. The flow images are also passed through a CNN and provided as input to the LSTM.  Flow CNN models have been shown to be beneficial for activity recognition \cite{simonyan14nips,donahue15cvpr}. 

To our knowledge, this is the first  approach to video description that uses a general sequence to sequence model.
This allows our model to (a) handle a variable number of input frames, (b) learn and use the temporal structure of the video and (c) learn a language model to generate natural, grammatical sentences. Our model is learned jointly and end-to-end,  incorporating both intensity and optical flow inputs, and does not require an explicit attention model. We demonstrate that S2VT achieves state-of-the-art performance on three diverse datasets, a standard YouTube corpus (MSVD) \cite{chen:acl11} and the M-VAD \cite{torabi15arxiv} and MPII Movie Description \cite{rohrbach15cvpr} datasets.
Our implementation (based on the \textit{Caffe}~\cite{caffe} deep learning framework) is available on github. 
{\small{\url{https://github.com/vsubhashini/caffe/tree/recurrent/examples/s2vt}}}.

%% file: related.tex
\section{Related Work}

Early work on video captioning considered tagging  videos with metadata \cite{aradhye09icdm} and clustering captions and videos \cite{huang2013multi,over12tv, wei2010multimodal} for retrieval tasks.
Several previous methods for generating sentence descriptions \cite{guadarrama13iccv,krishnamoorthy:aaai13,thomason14coling} used a two stage pipeline that first identifies the semantic content (subject, verb,  object) and then generates a sentence based on a template. This typically involved training individual classifiers  
to identify candidate objects, actions and scenes. They then use a probabilistic graphical model to combine 
the visual confidences with a language model
in order to estimate the most likely content (subject, verb, object, scene) in the video, which is then used to generate a sentence. While this simplified the problem by detaching content generation and surface realization, it requires selecting a set of relevant objects and actions to recognize. Moreover, a template-based approach to sentence generation is insufficient to model the richness of language used in human descriptions -- e.g., which attributes to use and how to combine them effectively to generate a good description. In contrast, our approach 
avoids the separation of content identification and sentence generation 
 by learning to directly map videos to full human-provided sentences, learning a language model simultaneously 
conditioned on visual features.

Our models take inspiration from the image caption generation models in
\cite{donahue15cvpr,vinyals14arxiv}. 
Their first step
is to generate a fixed length vector representation of an image by extracting features from a CNN. The next step  learns to decode this vector into a sequence of words composing the description of the image. While any RNN can be used in principle to decode the sequence, the resulting long-term dependencies can lead to inferior performance. To mitigate this issue, LSTM models have been exploited  as sequence decoders, as they are more suited to learning long-range dependencies. In addition, since we are using variable-length video as input,  
we use LSTMs as sequence to sequence transducers, following the language translation models of \cite{ilya:nips14}.

In \cite{venugopalan15naacl}, LSTMs are used to generate video descriptions
by pooling the representations of individual frames. Their technique extracts CNN features for frames in the video and then mean-pools the results to get a single feature vector representing the entire video. They then use an LSTM as a sequence decoder to generate a description based on this vector.
A major shortcoming of this approach is that this representation completely ignores the ordering of the video frames and fails to exploit any temporal information. 
The approach in \cite{donahue15cvpr} also generates video descriptions using an
LSTM; however,
they employ a version of the two-step approach that uses CRFs to
obtain semantic tuples of activity, object, tool, and locatation and then use an LSTM to
translate this tuple into a sentence.
Moreover, the model in \cite{donahue15cvpr} is applied
to the limited domain of cooking videos while ours is aimed at generating
descriptions for videos ``in the wild''.

Contemporaneous with our work, the approach in \cite{yao15arxiv} also addresses
the limitations of \cite{venugopalan15naacl} in two ways. First, they employ a
3-D convnet model that incorporates spatio-temporal motion features. To obtain
the features, they assume videos are of fixed volume 
{\it{(width, height, time)}}.
They extract dense trajectory features (HoG, HoF, MBH) \cite{wang2013action}
over non-overlapping cuboids and 
concatenate these to form the input. 
The 3-D convnet is pre-trained on video datasets for action recognition.
Second, they include an attention mechanism
that learns to weight the frame features non-uniformly conditioned on the previous word input(s) rather than uniformly weighting features from all frames as in \cite{venugopalan15naacl}. The 3-D convnet alone provides limited performance improvement, but in conjunction with the attention model it notably improves performance. We propose a simpler approach  to using temporal information 
by using an LSTM to encode the sequence of video frames into a distributed vector representation that is sufficient to  generate a sentential description.  Therefore, our direct sequence to sequence model does not require an explicit attention mechanism.

Another recent project \cite{nitish15arxiv} uses LSTMs to  predict the future frame sequence from an encoding of the previous frames. Their model is more similar to the language translation model in \cite{ilya:nips14}, which uses one LSTM to encode the input text into a fixed representation, and another LSTM to decode it into a different language. 
In contrast, we employ a single LSTM that learns both encoding and decoding based on the inputs it is provided. This allows the LSTM to share weights between encoding and decoding.
 
Other related work includes \cite{ng15arxiv,donahue15cvpr}, which uses LSTMs for activity classification, predicting an activity class for the representation of each image/flow frame. In contrast, our model generates captions after encoding the complete sequence of optical flow images. 
Specifically, our final model is an ensemble of the sequence to sequence models trained on raw images and optical flow images.

%% file: approach.tex
\section{Approach}

\begin{figure*}[!ht]
\begin{center}
\includegraphics[width=\linewidth]{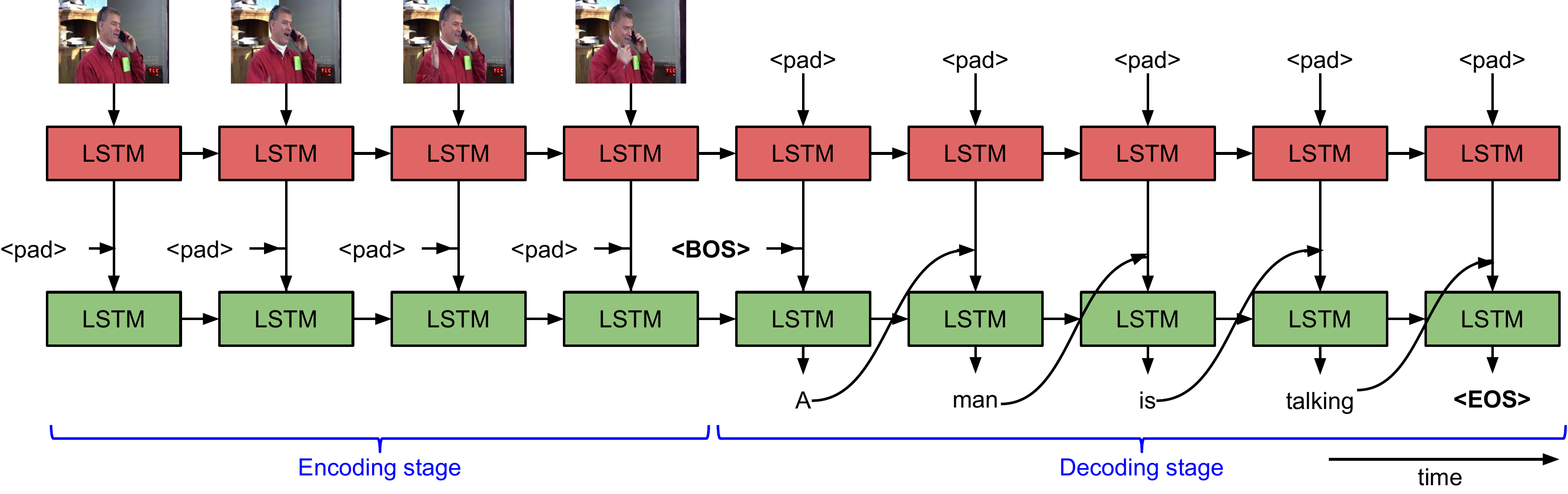}
\end{center}
 \caption{
We propose a stack of two LSTMs that learn a representation of a sequence of frames in order to decode it into a sentence that describes the event in the video. The top LSTM layer (colored red) models visual feature inputs. The second LSTM layer (colored green) models language given the text input and the hidden representation of the video sequence. We use $<$BOS$>$ to indicate begin-of-sentence and $<$EOS$>$ for the end-of-sentence tag. Zeros are used as a $<$pad$>$ when there is no input at the time step. }
\label{fig:approach}
\end{figure*}

We propose a sequence to sequence model for video description, where the input  is the sequence of video frames ($x_1, \ldots, x_n$), and the output is the sequence of words ($y_1, \ldots, y_m$). Naturally, both the input and output are of variable, potentially different, lengths.
In our case, there are typically many more frames than words.

In our model, we
estimate the conditional probability of an output sequence ($y_1, \ldots, y_m$) given an input sequence ($x_1, \ldots, x_n$) \ie
\vspace{-0.3cm}
\begin{equation}
\label{eq:condProb}
\vspace{-0.3cm}
p(y_1, \ldots, y_m | x_1, \ldots, x_n)
\end{equation}
This problem is analogous to machine translation between natural languages, where a sequence of words in the input language is translated to a sequence of words in the output language. Recently, \cite{cho2014properties,ilya:nips14} have shown how to effectively attack this sequence to sequence problem with an LSTM Recurrent Neural Network (RNN).
We extend this paradigm to inputs comprised of sequences of video frames, significantly simplifying prior RNN-based methods for video description. In the following, we describe our model and architecture in detail, as well as our input and output representation for video and sentences.

\subsection{LSTMs for sequence modeling} 
\label{subsec:lstm}

The main idea to handle variable-length input and output 
is  to first encode the input sequence of frames, one at a time, representing the video using a latent vector representation, and then decode from that representation to a sentence, one word at a time. 

Let us first recall the Long Short Term Memory RNN (LSTM),
originally proposed in \cite{schmidLSTM}. Relying on the LSTM unit proposed in \cite{zaremba2014learning}, for an input $x_t$ at time step $t$,
the LSTM computes a hidden/control state $h_t$ and a memory cell state $c_t$
which is an encoding of everything the cell has observed until time $t$: 
\noindent
\vspace{-0.2cm}
\begin{equation}\label{eqn:lstm}
\begin{array}{ccl}
i_t &=& \sigma(W_{xi}x_t + W_{hi}h_{t-1} + b_i) \\
f_t &=& \sigma(W_{xf}x_t + W_{hf}h_{t-1} + b_f) \\
o_t &=& \sigma(W_{xo}x_t + W_{ho}h_{t-1} + b_o) \\
g_t &=&   \phi(W_{xg}x_t + W_{hg}h_{t-1} + b_g) \\
c_t &=& f_t \odot c_{t-1} + i_t \odot g_t \\
h_t &=& o_t \odot \phi(c_t)
\end{array}
\vspace{-0.2cm}
\end{equation}
where $\sigma$ is the sigmoidal non-linearity, 
$\phi$ is the hyperbolic tangent non-linearity, $\odot$ represents the element-wise product
with the gate value, and the weight matrices denoted by $W_{ij}$ and biases $b_j$ are the trained parameters.

Thus, in the encoding phase, given an input sequence $X$ ($x_1, \ldots, x_n$),
the LSTM computes a sequence of hidden states ($h_1, \ldots, h_n$). During
decoding it defines a distribution over the output sequence $Y$ ($y_1, \ldots,
y_m$) given the input sequence $X$ as $p(Y|X)$ is
\vspace{-0.3cm}
\begin{align}\label{eqn:cond-prob}
\vspace{-0.3cm}
p(y_1, \ldots, y_m | x_1, \ldots, x_n) 
  = \prod_{t=1}^{m} p(y_t | h_{n+t-1}, y_{t-1})
\vspace{-0.2cm}
\end{align}
 where the distribution of $p(y_t | h_{n+t})$ is given by a
 {\it{softmax}} over all of the words in the vocabulary (see Equation
 \ref{eqn:softmax}). Note that $h_{n+t}$ is obtained from  $h_{n+t-1}, y_{t-1}$
 based on the recursion in Equation \ref{eqn:lstm}.

\subsection{Sequence to sequence video to text}
Our approach, \ApproachName, is depicted in Figure \ref{fig:approach}.
While \cite{cho2014properties,ilya:nips14} first encode the input sequence to a fixed length vector using one LSTM and then use another LSTM to map the vector to a sequence of outputs, we rely on a single LSTM for both the encoding and decoding stage. This allows parameter sharing between the encoding and decoding stage.

Our model uses a stack of two LSTMs with 1000 hidden units each. Figure \ref{fig:approach} shows the LSTM stack unrolled over time. When two LSTMs are stacked together, as in our case, 
the hidden representation ($h_t$) from the first LSTM layer (colored red)
is provided as the input ($x_t$) to the second LSTM (colored green).
 The top LSTM layer in our architecture is used to model the visual frame
 sequence, and the next layer is used to model the output word sequence. 

{\bf{Training and Inference}}
In the first several time steps, the top LSTM layer (colored red in Figure
\ref{fig:approach}) receives a sequence of frames and encodes them while the
second LSTM layer receives the hidden representation ($h_t$) and concatenates it
with null padded input words (zeros), which it then encodes. There is no loss
during this stage when the LSTMs are encoding. After all the frames in the video
clip are exhausted, the second LSTM layer is fed the beginning-of-sentence
($<$BOS$>$) tag, which prompts it to start decoding its current hidden
representation into a sequence of words. While training in the decoding stage, the
model maximizes for the log-likelihood of the predicted output sentence given
the hidden representation of the visual frame sequence, and the previous words
it has seen.
From Equation \ref{eqn:cond-prob} for a model with parameters $\theta$
and output sequence $Y = (y_1, \ldots,y_m)$, this is formulated as:
\vspace{-0.2cm}
\begin{equation}\label{eqn:loglikelihood}
 \theta^* = \argmax_{\theta} \text{ } \displaystyle\sum_{t=1}^{m} \text{log }p(y_t | h_{n+t-1}, y_{t-1}; \theta) 
 \end{equation}

This log-likelihood is optimized over the entire training dataset using stochastic gradient descent.
The loss is computed only when the LSTM is learning to decode. Since this loss
is propagated back in time, the LSTM learns to generate an appropriate hidden
state representation ($h_{n}$) of the input sequence. 
The output ($z_t$) of the second LSTM layer is used to obtain the emitted word ($y$). We apply a softmax
function to get the probability distribution over the words $y'$ in the vocabulary $V$:
\begin{equation} 
p(y|z_t) = \frac{\text{exp}(W_{y}z_{t})}{\sum_{y' \in
V} \text{exp}(W_{y'}z_{t})} \label{eqn:softmax}
\end{equation}
We note that, during the decoding phase, the visual frame representation for the first LSTM layer is simply a vector of zeros that acts as padding input. We require an explicit end-of-sentence tag ($<$EOS$>$) to terminate each sentence since this enables the model to define a distribution over sequences of varying lengths. At test time, during each decoding step we choose the word $y_t$ with the maximum probability after the softmax (from Equation \ref{eqn:softmax}) until it emits the $<$EOS$>$ token.

\subsection{Video and text representation} \label{sec:img_feats}
\noindent{\bf{RGB frames.}}
Similar to previous LSTM-based image captioning efforts \cite{donahue15cvpr,vinyals14arxiv} and video-to-text approaches \cite{venugopalan15naacl,yao15arxiv}, we apply a convolutional neural network (CNN) to input images and provide the output of the top layer as input to the LSTM unit. In this work, we report results using the output of the fc7 layer (after applying the ReLU non-linearity) on the Caffe Reference Net (a variant of AlexNet) and also the 16-layer VGG model \cite{vgg16arxiv}. 
We use CNNs that are pre-trained on the 1.2M image ILSVRC-2012 object classification subset of the ImageNet dataset \cite{imagenet2014} and made available publicly via the Caffe ModelZoo.\footnote{\url{https://github.com/BVLC/caffe/wiki/Model-Zoo}}
Each input video frame is scaled to 256x256, and is cropped to a random 227x227 region. It is then processed by the CNN. We remove the original last fully-connected classification layer and learn a new 
linear embedding of the features to a 500 dimensional space. The lower dimension features form the input ($x_t$) to the first LSTM layer.
The weights of the embedding are learned jointly with the LSTM layers during training.

\noindent{\bf{Optical Flow.}}
In addition to CNN outputs from raw image (RGB) frames, we also incorporate optical flow measures as input sequences to our architecture. Others \cite{ng15arxiv,donahue15cvpr} have shown that incorporating optical flow information to LSTMs improves activity classification. As many of our descriptions are activity centered, we explore this option for video description as well.
We follow the approach in \cite{donahue15cvpr,actiontubes}  and first extract classical variational optical flow features \cite{brox04flow}. We then create flow images (as seen in Figure \ref{fig:teaser}) in a manner similar to \cite{actiontubes}, by centering $x$ and $y$ flow values around 128 and multiplying by a scalar such that flow values fall between 0 and 255. We also calculate the flow magnitude and add it as a third channel to the flow image. We then use a CNN \cite{actiontubes} initialized with weights trained on the UCF101 video dataset to classify optical flow images into 101 activity classes. The fc6 layer activations of the CNN are embedded in a lower 500 dimensional space which is then given as input to the LSTM. The rest of the LSTM architecture remains unchanged for flow inputs.

In our combined model, we use a shallow fusion technique to integrate flow and RGB features. At each time step of the decoding phase, the model proposes a set of candidate words. We then rescore these hypotheses with the weighted sum of the scores by the flow and RGB networks, where we only need to recompute the score of each new word $p(y_t = y')$ as:
\vspace{-2.0mm}
$$ \alpha \cdot p_{rgb}(y_t = y') + (1 - \alpha) \cdot p_{flow} (y_t = y')
\vspace{-2.0mm}
$$
the hyper-parameter $\alpha$ is tuned on the validation set.

{\bf{Text input.}} The target output sequence of words are represented using
one-hot vector encoding (1-of-N coding, where N is the size of the vocabulary).
Similar to the treatment of frame features, we embed words to a lower
500 dimensional space by applying a linear transformation
to the input data and learning its parameters via back propagation.
The embedded word vector concatenated with the output ($h_t$) of the first
LSTM layer forms the input to the second LSTM layer (marked green in 
Figure \ref{fig:approach}). 
 When considering the output of the LSTM we apply a softmax over the complete vocabulary as in Equation \ref{eqn:softmax}.

%% file: evaluation.tex
\section{Experimental Setup}
This secction describes the evaluation of our approach. We first describe the datasets used, then the evaluation protocol, and then the details of our models.

\subsection{Video description datasets}
We report results on three video description corpora, namely the Microsoft Video
Description corpus (MSVD) \cite{chen:acl11}, the MPII Movie Description Corpus
(MPII-MD) \cite{rohrbach15cvpr}, and the Montreal Video Annotation Dataset
(M-VAD) \cite{torabi15arxiv}. Together they form the largest parallel corpora with
open domain video and natural language descriptions. While MSVD is based on web clips with short human-annotated sentences,  MPII-MD and M-VAD contain
Hollywood movie snippets with descriptions sourced from script data and audio
description. 
Statistics of each corpus are presented in Table
\ref{tab:corpus-stats}.
\begin{table}
\begin{center}
\begin{tabular}{lccc}
\toprule
  & MSVD & MPII-MD &  MVAD\\
 \cmidrule(lr){1-1}\cmidrule(lr){2-2}\cmidrule(lr){3-3}\cmidrule(lr){4-4}
\#-sentences & 80,827 & 68,375 & 56,634 \\
\#-tokens & 567,874 & 679,157 & 568,408 \\
vocab & 12,594 & 21,700 & 18,092 \\
\#-videos & 1,970 & 68,337 & 46,009 \\
avg. length & 10.2s & 3.9s & 6.2s \\
\#-sents per video & $\approx$41 & 1 & 1-2 \\
\bottomrule
\end{tabular}
\end{center}
\caption{Corpus Statistics. The the number of tokens in all datasets
are comparable, however MSVD has multiple descriptions for each video while the movie corpora (MPII-MD, MVAD) have a large number of clips with a
single description each. Thus, the number of video-sentence pairs in all
3 datasets are comparable.
}
\label{tab:corpus-stats}
\end{table}

\vspace{-0.3cm}
\subsubsection{Microsoft Video Description Corpus (MSVD)} 
The Microsoft Video description corpus \cite{chen:acl11}, is a collection of 
Youtube clips collected on Mechanical Turk by requesting workers to pick short
clips depicting a single activity. The videos were then used to elicit single sentence descriptions from annotators.
The original corpus has multi-lingual descriptions, in this work we use only the
English descriptions.
We do minimal
pre-processing on the text by converting all text to lower case, tokenizing the
sentences and removing punctuation.
We use the data splits provided by \cite{venugopalan15naacl}. Additionally,
in each video, we sample every tenth frame as done by \cite{venugopalan15naacl}.

\vspace{-0.3cm}
\subsubsection{MPII Movie Description Dataset (MPII-MD)}
MPII-MD \cite{rohrbach15cvpr} contains around 68,000 video clips extracted from 94 Hollywood movies. Each clip is accompanied with a single sentence description which is sourced from movie scripts and audio description (AD) data. AD or Descriptive Video Service (DVS) is an additional audio track that is added to the movies to describe explicit visual elements in a movie for the visually impaired.
Although the movie snippets are manually aligned to the descriptions, the data is very challenging due to the high diversity of visual and textual content, and the fact that most snippets have only a single reference sentence. We use the training/validation/test split provided by the authors and extract every fifth frame (videos are shorter than MSVD, averaging 94 frames).

\vspace{-0.3cm}
\subsubsection{Montreal Video Annotation Dataset (M-VAD)}
The M-VAD movie description corpus \cite{torabi15arxiv} is another recent collection of about 49,000 short video clips from 92 movies. 
It is similar to MPII-MD, but contains only AD data with automatic alignment. We use the same setup as for MPII-MD.

\subsection{Evaluation Metrics}
Quantitative evaluation of the models are performed using the METEOR \cite{banerjee2005meteor} metric which was originally proposed to evaluate machine translation results. The METEOR score is computed based on the alignment between a given hypothesis sentence and a set of candidate reference sentences. 
METEOR compares exact token matches, stemmed tokens, paraphrase matches, as well as semantically similar matches using WordNet synonyms. This semantic aspect of METEOR distinguishes it from others such as BLEU~\cite{papineni:2002}, ROUGE-L~\cite{lin2004rouge}, or CIDEr~\cite{cider}.
The authors of CIDEr~\cite{cider} evaluated these four measures for image description. They showed that METEOR is always better than BLEU and ROUGE and outperforms CIDEr when the number of references are small (CIDEr is comparable to METEOR when the number of references are large). Since MPII-MD and M-VAD have only a single reference, we decided to use METEOR in all our evaluations.
We employ METEOR version 1.5 \footnote{\url{http://www.cs.cmu.edu/~alavie/METEOR}} using the code\footnote{\url{https://github.com/tylin/coco-caption}} released with the Microsoft COCO Evaluation Server \cite{chen15arxiv}.

\subsection{Experimental details of our models}
All our models take as input either the raw RGB frames directly feeding into the CNN, or pre-processed optical flow images (described in Section \ref{sec:img_feats}).
In all of our models, we unroll the LSTM to a fixed 80 time steps during training. We found this to be a good trade-off between memory consumption and
the ability to provide many frames (videos) to the LSTM.
This setting allows us to fit multiple videos in a single mini-batch (up to 8 for AlexNet and up to 3 for flow models).
We note that 94$\%$ of the YouTube training videos satisfied this limit (with frames sampled at the rate of 1 in 10). 
For videos with fewer than 80 time steps (of words and frames), we pad the remaining inputs with zeros. For longer videos, we truncate the number of frames to ensure that the sum of the number of frames and words is within this limit.
At test time, we do not constrain the length of the video and our model views all sampled frames. We use the pre-trained AlexNet and VGG CNNs. For VGG, we fix all layers below fc7 to reduce memory consumption and allow faster training.

We compare our sequence to sequence LSTM architecture with RGB image features
extracted from both AlexNet, and the 16-layer VGG network. In order to compare
features from the VGG network with previous models, we include the performance
of the mean-pooled model proposed in \cite{venugopalan15naacl} using the output of
the fc7 layer from the 16 layer VGG as a baseline (line 3,
\Tableref{tab:bleumeteor}). All our sequence to sequence
models are referenced in Table \ref{tab:bleumeteor} under S2VT. Our first variant,
RGB (AlexNet) is the end-to-end model that uses AlexNet on RGB frames.
Flow (AlexNet) refers to the model that is obtained by training on optical flow images.
RGB (VGG) refers to the model with the 16-layer VGG model on RGB image frames. We also experiment with randomly re-ordered input frames  (line 10)
to verify that S2VT learns temporal-sequence information.
Our final model is an ensemble of the RGB (VGG) and Flow (AlexNet) where the
prediction at each time step is a weighted average of the prediction from the individual models.

\subsection{Related approaches}
We compare our sequence to sequence models against the 
factor graph model (FGM)  in \cite{thomason14coling}, the mean-pooled models in
\cite{venugopalan15naacl} and the Soft-Attention models of \cite{yao15arxiv}.\\
{\bf{FGM}} proposed in \cite{thomason14coling} uses a two step approach to first obtain confidences on subject, verb, object and scene (S,V,O,P) elements and combines these with confidences from a language model using a factor graph to infer the most likely (S,V,O,P) tuple in the video. It then generates a sentence based on a template.\\
The {\bf{Mean Pool}} model proposed in \cite{venugopalan15naacl} pools AlexNet fc7 activations across all frames to create a fixed-length vector representation of the video. It then uses an LSTM to then decode the vector into a sequence of words. Further, the model ia pre-trained on the Flickr30k \cite{flickr30k} and MSCOCO \cite{coco2014} image-caption datasets and fine-tuned on MSVD for a significant improvement in performance. We compare our models against their basic mean-pooled model and their best model obtained from fine-tuning on Flickr30k and COCO datasets.
We also compare against the GoogleNet \cite{googlenet} variant of the mean-pooled model reported in \cite{yao15arxiv}.\\
The {\bf{Temporal-Attention}} model in \cite{yao15arxiv} is a combination of weighted attention over a fixed set of video frames with input features from GoogleNet and a 3D-convnet trained on HoG, HoF and MBH features from an activity classification model.

%% file: discussion.tex
\section{Results and Discussion}
This section discussses the result of our evaluation shown in \Tablesref{tab:bleumeteor}, \ref{tab:results:md}, and \ref{tab:results:mvad}.

\input{result-table-MSVD}

\subsection{MSVD dataset}
\Tableref{tab:bleumeteor} shows the results on the MSVD dataset. Rows 1 through 7 present related approaches and the rest are variants of our S2VT approach. 
Our basic S2VT AlexNet model on RGB video frames (line 9 in \Tableref{tab:bleumeteor}) achieves 27.9\% METEOR and improves over the basic mean-pooled model in \cite{venugopalan15naacl} (line 2, 26.9\%) as well as the VGG  mean-pooled model (line 3, 27.7\%);suggesting that S2VT is a more powerful approach.
When the model is trained with 
randomly-ordered frames (line 10 in \Tableref{tab:bleumeteor}), the score is
considerably lower, clearly demonstrating that the model benefits from exploiting temporal structure.

Our S2VT model which uses flow images (line 8) achieves only 24.3\% METEOR but improves the performance of our VGG model from 29.2\% (line 11) to 29.8\% (line 12), when combined. 
A reason for the low performance of the flow model could be that optical flow features even for the same activity can vary significantly with context e.g. `panda eating' vs `person eating'. Also, the model only receives very weak signals with regard to the kind of activities depicted in YouTube videos. Some commonly used verbs such as ``play'' are polysemous and can refer to playing a musical instrument (``playing a guitar'') or playing a sport (``playing golf''). However, integrating RGB with Flow improves the quality of descriptions.

Our ensemble using both RGB and Flow performs slightly better than the best model proposed in {\cite{yao15arxiv}}, temporal attention with GoogleNet + 3D-CNN (line 7). The modest size of the improvement is likely due to the much stronger 3D-CNN features (as the difference to GoogleNet alone (line 6) suggests).
Thus, the closest comparison between the Temporal Attention Model \cite{yao15arxiv}
and S2VT is arguably S2VT with VGG (line 12) vs. their GoogleNet-only model
(line 6).

Figure \ref{fig:msvd-cherry-lemon} shows
descriptions generated by our model on sample Youtube clips from MSVD.
To compare the originality in generation, we
compute the Levenshtein distance of the predicted sentences with those in the
training set. From \Tableref{tab:edit-dist}, for the MSVD corpus, 42.9\% of the predictions 
are identical to some training sentence, and
another 38.3\% can be obtained by inserting, deleting or substituting one word from
some sentence in the training corpus.
We note that many of the descriptions generated are relevant.
\begin{table}
\begin{center}
\begin{tabular}{lcccc}
\toprule
Edit-Distance & $k=0$ & $k<=1$ & $k<=2$ & $k<=3$\\ 
\cmidrule(lr){1-1}\cmidrule(lr){2-5}
MSVD  & 42.9 & 81.2 & 93.6 & 96.6 \\ 
MPII-MD  & 28.8 & 43.5 & 56.4 & 83.0 \\ 
MVAD  & 15.6 & 28.7 & 37.8 & 45.0 \\ 
\bottomrule
\end{tabular}
\end{center}
\caption{
Percentage of generated sentences which match a sentence of the training set
with an edit (Levenshtein) distance of less than 4.
All values reported in percentage (\%).
}
\label{tab:edit-dist}
\end{table}

\subsection{Movie description datasets}

For the more challenging MPII-MD and M-VAD datasets we use our single best model, namely S2VT trained on RGB frames and VGG.
To avoid over-fitting on the movie corpora we employ drop-out which has proved to be beneficial on these datasets \cite{rohrbach15gcpr}. We found it was best to use dropout at the inputs and outputs of both LSTM layers. Further, we used ADAM \cite{kingma14adam} for optimization with a first momentum coefficient of
0.9 and a second momentum coefficient of 0.999.
For MPII-MD, reported in \Tableref{tab:results:md}, we improve  over the SMT approach from \cite{rohrbach15cvpr} from 5.6\% to 7.1\% METEOR and over Mean pooling \cite{venugopalan15naacl} by 0.4\%. Our performance is similar to Visual-Labels \cite{rohrbach15gcpr}, a contemporaneous LSTM-based approach which uses no temporal encoding, but more diverse visual features, namely object detectors, as well as activity and scene classifiers.

On M-VAD we achieve 6.7\% METEOR which significantly outperforms the temporal attention model \cite{yao15arxiv} (4.3\%)\footnote{\label{fn:resultcomputation}We report results using the predictions provided by \cite{yao15arxiv} but using the orginal COCO Evaluation scripts.
\cite{yao15arxiv} report 5.7\% METEOR for their temporal attention + 3D-CNN model using a different tokenization.} and Mean pooling (6.1\%). On this dataset we also outperform Visual-Labels \cite{rohrbach15gcpr} (6.3\%).

We report results on the \textbf{LSMDC} challenge\footnote{LSMDC: {\url{sites.google.com/site/describingmovies}}}, which combines M-VAD and MPII-MD. S2VT achieves 7.0\% METEOR on the public test set using the evaluation server.
 
 In Figure \ref{fig:mvad-iccv} we present descriptions generated by our model on some sample clips from the M-VAD dataset. More example video clips,  generated sentences, and data are available on the authors' webpages\footnote{\url{http://vsubhashini.github.io/s2vt.html}}.

\begin{table}
\begin{center}
\begin{tabular}{lc}
\toprule
Approach & METEOR \\
\cmidrule(lr){1-1}\cmidrule(lr){2-2}
SMT (best variant) \cite{rohrbach15cvpr} & 5.6 \\
Visual-Labels \cite{rohrbach15gcpr} & 7.0 \\
Mean pool (VGG) & 6.7\\ 
S2VT: RGB (VGG), ours & 7.1\\ 
\bottomrule
\end{tabular}
\end{center}
\caption{
MPII-MD dataset (METEOR in \%, higher is better).
}
\label{tab:results:md}
\end{table}

\begin{table}
\begin{center}
\begin{tabular}{lc}
\toprule
Approach & METEOR \\
\cmidrule(lr){1-1}\cmidrule(lr){2-2}
Visual-Labels \cite{rohrbach15gcpr} & 6.3 \\
Temporal att. (GoogleNet+3D-CNN) \cite{yao15arxiv}\textsuperscript{
\ref{fn:resultcomputation}}& 4.3 \\
Mean pool (VGG) & 6.1 \\ 
S2VT: RGB (VGG), ours & 6.7 \\ 
\bottomrule
\end{tabular}
\end{center}
\caption{
M-VAD dataset (METEOR in \%, higher is better). 
}
\label{tab:results:mvad}
\end{table}

%% file: result-table-MSVD.tex
\begin{table}
\begin{center}
\begin{tabular}{lcl@{}r}
\toprule
Model & \multicolumn{1}{r}{METEOR} & \\
\cmidrule(lr){1-1}\cmidrule(lr){2-2}
FGM {\cite{thomason14coling}}& 23.9 & {\small\tt{(1)}}\\ 
Mean pool \\
\ - AlexNet {\cite{venugopalan15naacl}} & 26.9 & {\small\tt{(2)}}\\ 
\ - VGG   & 27.7 & {\small\tt{(3)}}\\ 
\ - AlexNet COCO pre-trained  {\cite{venugopalan15naacl}} & 29.1 & {\small\tt{(4)}}\\ 
\ - GoogleNet \cite{yao15arxiv} & 28.7 & {\small\tt{(5)}}\\
Temporal attention\\ 
\ - GoogleNet {\cite{yao15arxiv}}  & 29.0 & {\small\tt{(6)}}\\
\ - GoogleNet + 3D-CNN {\cite{yao15arxiv}}  & 29.6 & {\small\tt{(7)}}\\ 
\cmidrule(lr){1-1}\cmidrule(lr){2-2}
S2VT (ours)\\
\ - Flow (AlexNet) & 24.3 & {\small\tt{(8)}}\\ 
\ - RGB (AlexNet) & 27.9 & {\small\tt{(9)}}\\ 
\ - RGB (VGG) random frame order & 28.2 &{\small\tt{(10)}}\\ 
\ - RGB (VGG) & 29.2 & {\small\tt{(11)}}\\ 
\ - RGB (VGG) +  Flow (AlexNet) &  29.8 & {\small\tt{(12)}}\\
\bottomrule
\end{tabular}
\end{center}
\caption{MSVD dataset (METEOR in \%, higher is better).}
\label{tab:bleumeteor}
\end{table}

%% file: conclusion.tex
\section{Conclusion}
This paper proposed a novel approach to video description. In contrast to related work, we construct descriptions using a sequence to sequence model, where frames are first read sequentially and then words are generated sequentially. This allows us to handle variable-length input and output while simultaneously modeling temporal structure.
Our model achieves state-of-the-art performance on the MSVD dataset, and outperforms related work on two large and challenging movie-description datasets. Despite its conceptual simplicity, our model significantly benefits from additional data, suggesting that it has a high model capacity, and is able to learn complex temporal structure in the input and output sequences for challenging movie-description datasets.

\begin{figure*}[!htb]
\begin{center}
\vspace{-0.3cm}
\includegraphics[width=\linewidth]{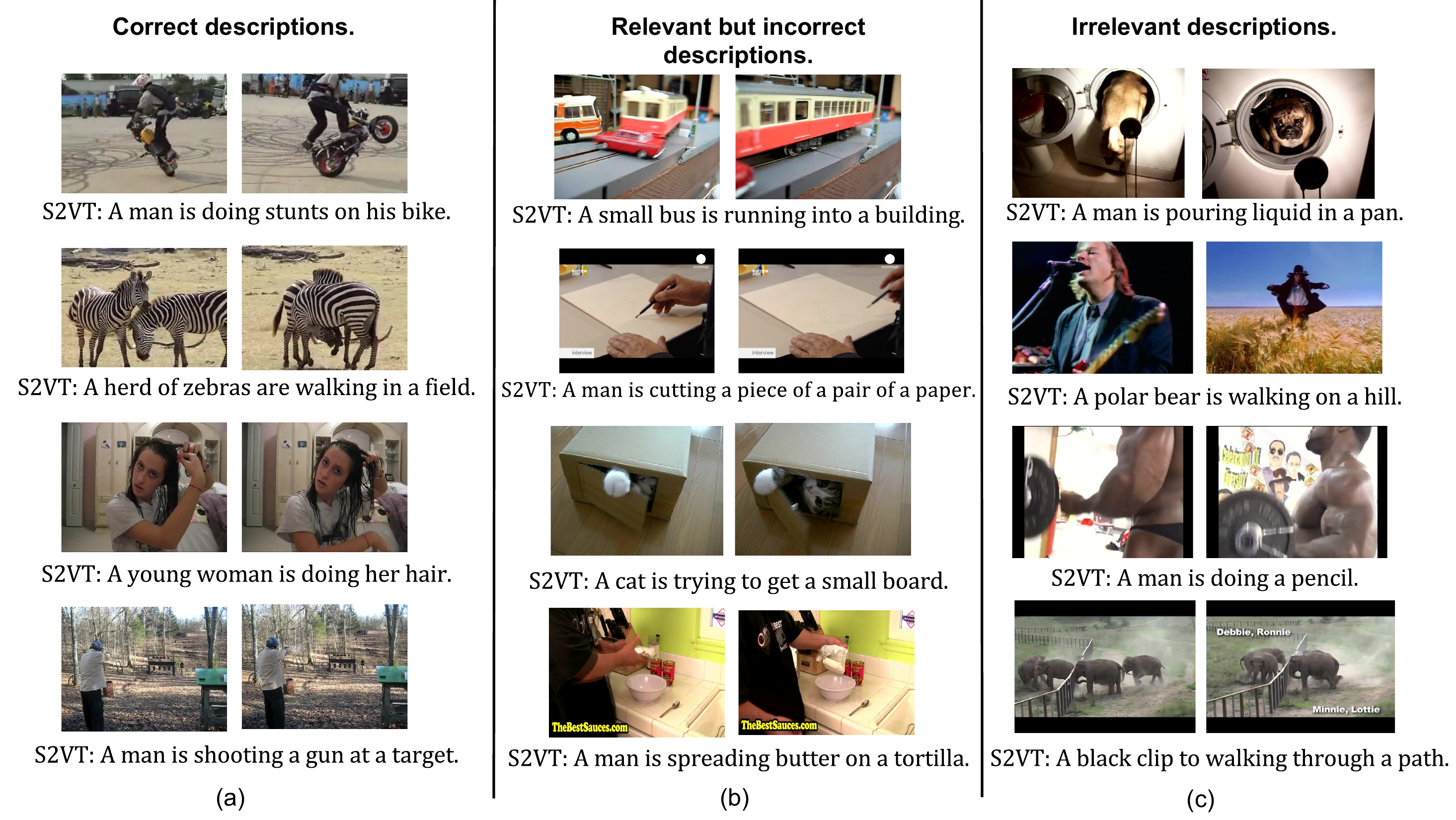}
\end{center}
\vspace{-0.5cm}
 \caption{Qualitative results on MSVD YouTube dataset from our S2VT model
 (RGB on VGG net). (a) Correct descriptions involving different objects and actions for several videos.  (b)  Relevant but incorrect descriptions.  (c) Descriptions that are irrelevant to the event in the video.
 }
\label{fig:msvd-cherry-lemon}
\end{figure*}

\begin{figure*}[!htb]
\begin{center}
\vspace{-0.2cm}
\includegraphics[width=\linewidth]{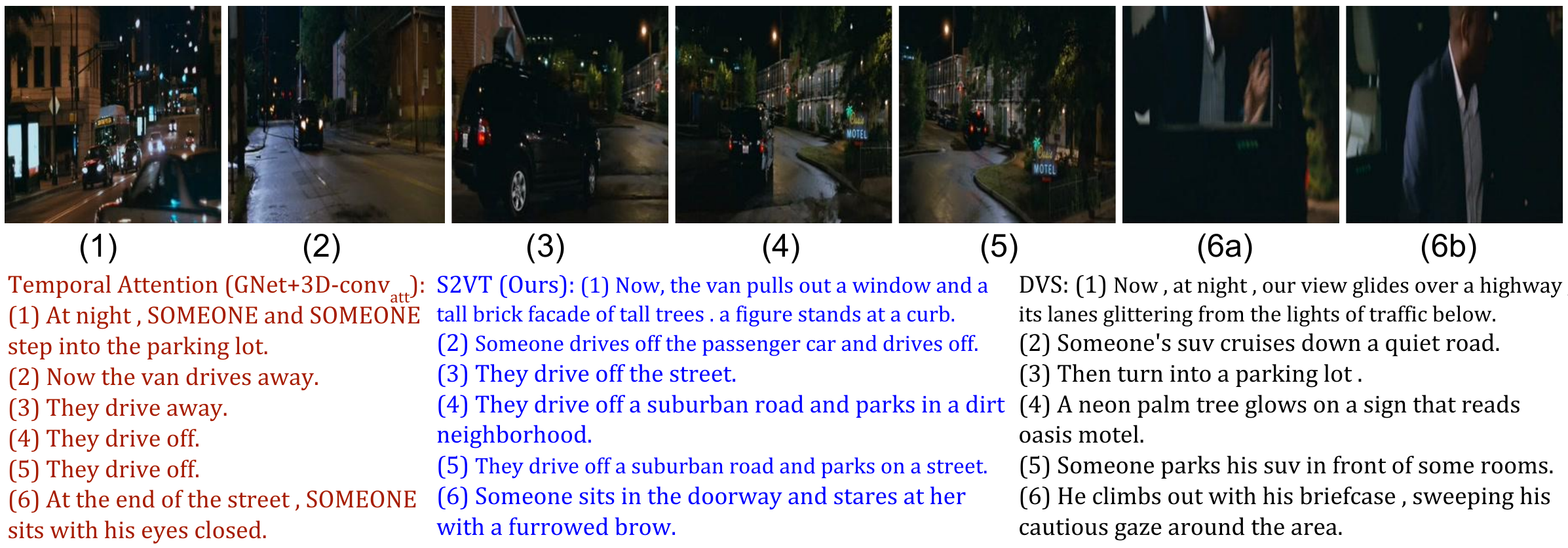}
\end{center}
\vspace{-0.2cm}
 \caption{M-VAD Movie corpus: Representative frame from 6 contiguous
 clips from the movie ``Big Mommas: Like Father, Like Son''. From left: Temporal Attention (GoogleNet+3D-CNN) \cite{yao15arxiv}, S2VT (in blue) trained on the M-VAD dataset, and DVS: ground truth.
 }
\label{fig:mvad-iccv}
\end{figure*}

%% file: ack.tex
\section*{Acknowledgments}
We thank Lisa Anne Hendricks, Matthew Hausknecht, Damian Mrowca for
helpful discussions; and Anna Rohrbach for help with both movie corpora; and the anonymous reviewers for insightful comments and suggestions.
We acknowledge support from ONR ATL Grant N00014-11-1-010,
DARPA, AFRL, DoD MURI award N000141110688, DEFT program (AFRL grant FA8750-13-2-0026), NSF awards
IIS-1427425, IIS-1451244, and IIS-1212798,
and Berkeley Vision and Learning Center. Raymond and Kate acknowledge support from Google. Marcus was supported by the FITweltweit-Program of the German Academic Exchange Service (DAAD).